\documentclass[sigconf]{acmart}
\AtBeginDocument{%
  }

\copyrightyear{2026}
\acmYear{2026}
\setcopyright{cc}
\setcctype{by}
\acmConference[WWW '26]{Proceedings of the ACM Web Conference 2026}{April 13--17, 2026}{Dubai, United Arab Emirates}
\acmBooktitle{Proceedings of the ACM Web Conference 2026 (WWW '26), April 13--17, 2026, Dubai, United Arab Emirates}
\acmPrice{}
\acmDOI{10.1145/3774904.3792811}
\acmISBN{979-8-4007-2307-0/2026/04}

\usepackage[utf8]{inputenc}
\usepackage[T1]{fontenc}
\usepackage{microtype}
\usepackage{enumitem}
\usepackage{multirow}

\hypersetup{colorlinks=true,linkcolor=black,citecolor=black,urlcolor=blue}

\newcommand{\R}{\mathbb{R}}

\DeclareMathOperator*{\softmax}{softmax}

\begin{document}

\title{Make It Long, Keep It Fast: End-to-End 10k-Sequence Modeling \\
at Billion Scale on Douyin Recommendation}

\author{Lin Guan}
\authornote{Contributed equally.}
\email{guanlin.13@gmail.com}
\orcid{0009-0006-6034-4819}
\affiliation{%
  \institution{ByteDance}
  \city{Beijing}
  \country{China}
}

\author{Jia-Qi Yang}
\authornotemark[1]
\email{yangjiaqi.yjq@bytedance.com}
\orcid{0000-0002-6331-0829}
\affiliation{%
  \institution{ByteDance}
  \city{Shanghai}
  \country{China}
}

\author{Zhishan Zhao}
\authornotemark[1]
\email{zhaozhishan@bytedance.com}
\orcid{0009-0000-4023-2197}
\affiliation{%
  \institution{ByteDance}
  \city{Beijing}
  \country{China}
}

\author{Beichuan Zhang}
\authornotemark[1]
\email{zhangbeichuan.123@bytedance.com}
\orcid{0000-0003-1163-4036}
\affiliation{%
  \institution{ByteDance}
  \city{Beijing}
  \country{China}
}

\author{Bo Sun}
\email{zhangyang.2583@bytedance.com}
\orcid{0009-0004-0366-1047}
\affiliation{%
  \institution{ByteDance}
  \city{San Jose}
  \state{CA}
  \country{USA}
}

\author{Xuanyuan Luo}
\email{xuanyuanluo@bytedance.com}
\orcid{0009-0008-8412-3222}
\affiliation{%
  \institution{ByteDance}
  \city{Hangzhou}
  \state{Zhejiang}
  \country{China}
}

\author{Jinan Ni}
\email{nijinan@bytedance.com}
\orcid{0009-0004-3287-4770}
\affiliation{%
  \institution{ByteDance}
  \city{Shanghai}
  \country{China}
}

\author{Xiaowen Li}
\email{lixiaowen.911@bytedance.com}
\orcid{0009-0000-9688-8339}
\affiliation{%
  \institution{ByteDance}
  \city{Beijing}
  \country{China}
}

\author{Yuhang Qi}
\email{qiyuhang@bytedance.com}
\orcid{0009-0008-7659-3160}
\affiliation{%
  \institution{ByteDance}
  \city{Hangzhou}
  \state{Zhejiang}
  \country{China}
}

\author{Zhifang Fan}
\email{fanzhifangfzf@gmail.com}
\orcid{0009-0003-5234-3224}
\affiliation{%
  \institution{ByteDance}
  \city{Shanghai}
  \country{China}
}

\author{Hangyu Wang}
\email{wanghangyu.123@bytedance.com}
\orcid{0009-0006-0410-5426}
\affiliation{%
  \institution{ByteDance}
  \city{Shanghai}
  \country{China}
}

\author{Qiwei Chen}
\authornote{Corresponding author.}
\email{chenqiwei05@gmail.com}
\orcid{0009-0002-8920-1716}
\affiliation{%
  \institution{ByteDance}
  \city{Shanghai}
  \country{China}
}

\author{Yi Cheng}
\email{chengyi.23@bytedance.com}
\orcid{0009-0006-2950-1548}
\affiliation{%
  \institution{ByteDance}
  \city{Beijing}
  \country{China}
}

\author{Feng Zhang}
\email{feng.zhang@bytedance.com}
\orcid{0009-0009-2635-2710}
\affiliation{%
  \institution{ByteDance}
  \city{Shanghai}
  \country{China}
}

\author{Xiao Yang}
\email{wuqi.shaw@bytedance.com}
\orcid{0000-0003-0712-0550}
\affiliation{%
  \institution{ByteDance}
  \city{Beijing}
  \country{China}
}

\renewcommand{\shortauthors}{Lin Guan et al.}
\renewcommand{\shorttitle}{End-to-End 10k-Sequence Modeling at Billion Scale on Douyin}

\begin{abstract}
Short-video recommenders such as Douyin must exploit extremely long user histories without breaking latency or cost budgets. We present an end-to-end system that scales long-sequence modeling to \emph{10k}-length histories in production. First, we introduce \emph{Stacked Target-to-History Cross Attention (STCA)}, which replaces history self-attention with stacked cross-attention from the target to the history, reducing complexity from quadratic to \emph{linear} in sequence length and enabling efficient end-to-end training. Second, we propose \emph{Request Level Batching (RLB)}, a user-centric batching scheme that aggregates multiple targets for the same user/request to share the user-side encoding, substantially lowering sequence-related storage, communication, and compute without changing the learning objective. Third, we design a \emph{length-extrapolative} training strategy---train on shorter windows, infer on much longer ones---so the model generalizes to \emph{10k} histories without additional training cost. Across offline and online experiments, we observe predictable, monotonic gains as we scale history length and model capacity, mirroring the \emph{scaling law} behavior observed in large language models. Deployed at full traffic on Douyin, our system delivers significant improvements on key engagement metrics while meeting production latency, demonstrating a practical path to scaling end-to-end long-sequence recommendation to the 10k regime.
\end{abstract}



\maketitle

\section{Introduction}
Deep neural networks have become the backbone of modern recommender systems, powering applications in e-commerce, news feeds, and short-video platforms\cite{DBLP:journals/csur/ZhangYST19,DBLP:conf/recsys/Cheng0HSCAACCIA16,DBLP:conf/recsys/CovingtonAS16,DBLP:conf/www/HeLZNHC17}. A key reason is their ability to leverage user behavior sequences, as past interactions provide essential signals for inferring preferences\cite{DBLP:journals/corr/HidasiKBT15,DBLP:conf/cikm/LiLWXZHKCLL19}. In short-video recommendation such as Douyin\cite{DBLP:journals/corr/abs-2507-15551}, user histories are often thousands of videos long. If effectively utilized, these long sequences can substantially improve ranking performance\cite{DBLP:conf/icdm/KangM18,DBLP:conf/cikm/SunLWPLOJ19,DBLP:conf/kdd/ZhouZSFZMYJLG18}.

\begin{figure}[!t]
  \centering
  \includegraphics[width=\linewidth]{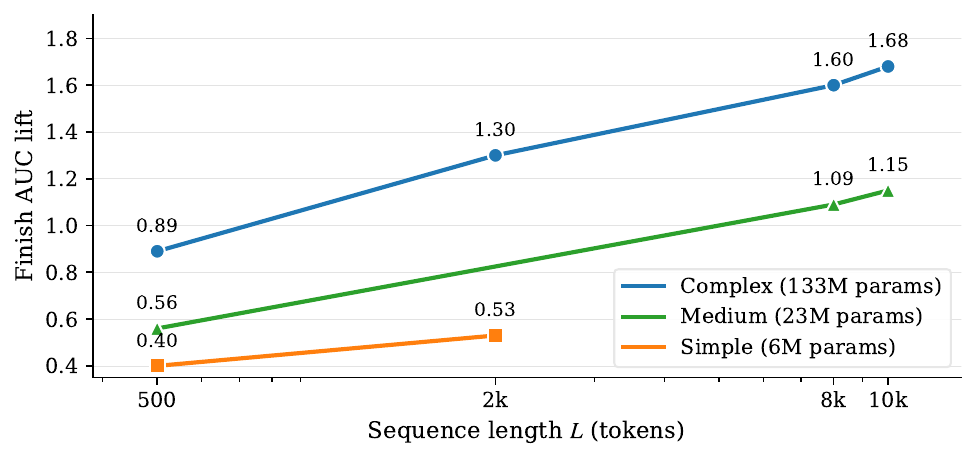}
  \caption{\textbf{Scaling with sequence length and model capacity.}
  Finish AUC lift (percentage points) for the sequence module (STCA) as we increase user sequence length ($500\!\rightarrow\!10\mathrm{k}$ tokens) and sequence-module capacity (Simple: 6M, Medium: 23M, Complex: 133M parameters).\label{fig:intro-scaling}}
\end{figure}

The importance of modeling longer sequences relates to the scaling law of deep learning: performance tends to improve predictably with more data, parameters, and compute\cite{DBLP:journals/corr/abs-2001-08361,DBLP:journals/corr/abs-2203-15556}. Unlike NLP/CV where scaling often comes from enlarging datasets, recommendation is constrained by user-generated data\cite{DBLP:conf/uai/MarlinZRS07,DBLP:conf/icml/SchnabelSSCJ16}. A natural way to expose more information is to use longer histories\cite{DBLP:journals/corr/abs-1905-06874,DBLP:conf/cikm/LinZWDCW22,DBLP:conf/ijcai/0001LGQZ0T23}. However, most systems adopt a two-stage paradigm\cite{DBLP:conf/cikm/PiZZWRFZG20,DBLP:conf/sigir/Qin0WJF020,DBLP:conf/kdd/ChangZFZGLHLNSG23,DBLP:conf/cikm/SiGSZLHCYZLZZN024}: retrieve a small set similar to the target and feed the truncated sequence to the ranker. While efficient, this breaks end-to-end optimization and discards valuable interactions. \emph{Our empirical results (Fig.~\ref{fig:intro-scaling}) show that when the architecture and system support true end-to-end long-sequence training and inference, model quality scales smoothly with both sequence length and sequence-module capacity, echoing scaling-law behavior in other modalities.}

To truly unlock scaling for large recommendation models, end-to-end training over long sequences must coexist with strict online latency and cost budgets\cite{DBLP:journals/corr/abs-2506-02267,DBLP:conf/sigmod/KersbergenSS22}. Designs must allocate computation selectively, and longer histories amplify storage, communication, and compute in distributed training\cite{DBLP:conf/mlsys/LuoZTMCCLHZLWPM24,DBLP:conf/isca/MudigereHHJT0LO22,DBLP:journals/corr/abs-2003-09518}. To this end, we combine: (i) a \emph{target-centric}, single-query cross-attention model (STCA) with per-layer cost linear in sequence length; (ii) \emph{Request Level Batching (RLB)}, which amortizes user-side encoding across multiple targets within a request and \emph{can be extended to share across multiple requests for the same user/session}, offering further efficiency gains; and (iii) a ``train sparsely/infer densely'' regimen: we train on sequences with an \emph{average length of around 2k} tokens, yet \emph{extrapolate to 10k} at serving, preserving end-to-end modeling without increasing training compute. Together these components enable predictable gains with increasing length and capacity, \emph{consistent with scaling-law behavior} (Figure~\ref{fig:intro-scaling}).

\paragraph{Our contributions.}
We make three contributions toward practical end-to-end long-sequence recommendation:
\begin{itemize}[leftmargin=1.25em]
\item \textbf{Sequence-efficient architecture.} We propose \emph{Stacked Target-to-History Cross Attention} (STCA), which prioritizes cross-attention between the target item and the history while omitting history self-attention, reducing complexity from $O(L^2)$ to $O(L)$ in sequence length $L$. Stacking multiple layers captures higher-order dependencies via target-conditioned fusion.
\item \textbf{Scalable training via user-centric batching.} We introduce \emph{Request Level Batching (RLB)}, aggregating samples from the same user to share one user-side encoding across multiple targets. Beyond a single request, RLB naturally extends to \emph{multi-request} sharing for the same user/session, further cutting memory, communication, and computation (up to $8\times$ reduction reported under request-level sharing) while remaining an unbiased estimator of empirical risk.
\item \textbf{Train sparsely, infer densely.} We adopt a length-extrapolative regimen that trains on sequences with an average length of $\sim$2k but serves on sequences up to 10k, decoupling training cost from deployment-time context length and realizing long-sequence gains without additional training compute.
\end{itemize}
These innovations provide a practical framework for scaling along the sequence dimension under production constraints. Deployed in a large-scale short-video platform, our methods deliver significant online gains across multiple business metrics.

\section{Background and Notations}
\label{sec:background}

\paragraph{Notation.}
We use the following symbols throughout. The user interaction history is
$\mathcal{H}=\{(v_i,a_i)\}_{i=1}^L$ of length $L$, where $v_i$ is the feature
vector of the $i$-th historical video and $a_i$ is its interaction type; $t$
denotes the candidate (target) video to be ranked. Embeddings are
$\mathbf{x}_i$ for $(v_i,a_i)$ and $\mathbf{x}_t$ for $t$. Let $d$ be the
embedding dimensionality, $r$ the expansion ratio in the \emph{SwiGLUFFN} width,
$h$ the number of attention heads, and $M$ the number of stacked cross-attention
layers. The model outputs $\hat{y}\in[0,1]$, the predicted finish probability
for $t$, and $y\in\{0,1\}$ is the ground-truth finish label.
We refer to each historical interaction $(v_i,a_i)$ as a \emph{token}; thus the sequence length $L$ denotes the number of history interactions (tokens).

\paragraph{Task.}
We consider large-scale short-video recommendation (e.g., Douyin / TikTok),
where the system ranks a set of candidate videos for a user. In practice, the
final score may combine multiple objectives (finish, click, etc.); for clarity
we focus on predicting the \emph{finish rate}. Given input $(\mathcal{H}, t)$,
the ranking model outputs a scalar $\hat{y}\in[0,1]$, the estimated probability
that the user finishes $t$ conditioned on $\mathcal{H}$. This notation will be
used throughout when introducing the model and training strategies.

\section{Method}\label{sec:method}

\begin{figure*}[t]
  \centering
  \includegraphics[width=\linewidth]{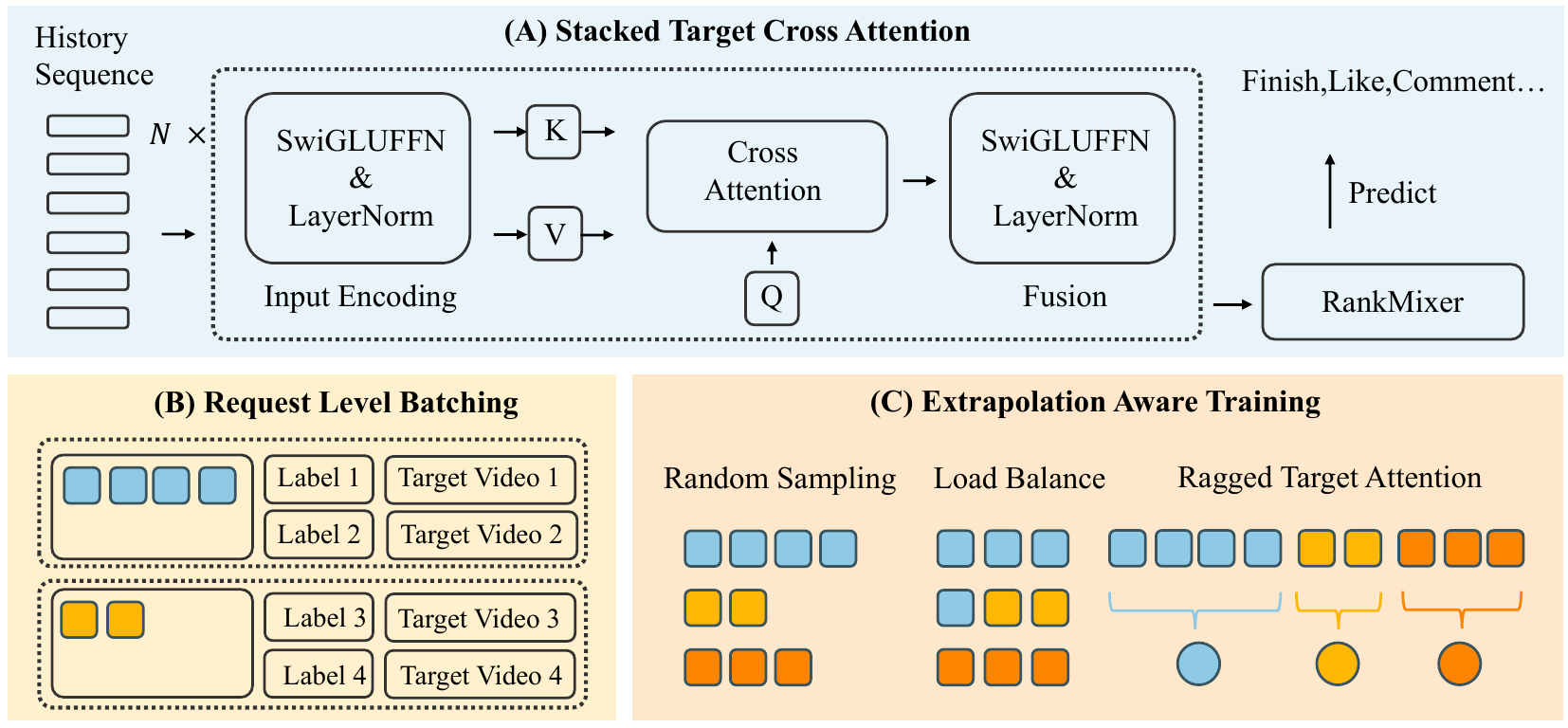}
  \caption{\textbf{Overview of our long-history ranking stack.}
  \textbf{(A) Stacked Target Cross Attention:} single-query cross attention from the target to the full history with layer-wise fusion, enabling linear scaling in history length and end-to-end optimization.
  \textbf{(B) Request Level Batching:} compute the user/history encoding once per request and reuse it across multiple targets to reduce bandwidth and compute.
  \textbf{(C) Extrapolation Aware Training:} sample shorter histories during training and serve longer histories at inference (Train Sparsely, Infer Densely).}
  \label{fig:struct-intro}
\end{figure*}

We develop an end-to-end framework for long-sequence modeling in short-video ranking that couples a sequence-efficient architecture, a user-centric batching strategy, and a length-extrapolative training regimen as depicted in Figure \ref{fig:struct-intro}.

\subsection{Stacked Target Cross Attention (STCA)}

In ranking, the principal signal for predicting the user’s response to $t$ arises from \emph{direct} interactions between $t$ and the user’s historical behaviors, whereas second-order relations among historical items are comparatively less informative. Yet Transformer-style self-attention over $[t;\mathcal{H}]$ incurs $O(L^2)$ cost in the history length and therefore constrains $L$.

We make an explicit capacity–cost trade-off: we \emph{de-emphasize} explicit history–history interactions and instead use a \emph{single-query} \emph{target-to-history cross attention} (STCA). By treating the target as the sole query, the per-layer complexity becomes linear in $L$ ($O(Ldh)$), focusing compute exactly on target--history relevance. In addition, with a single query we can reorder the computation (Sec.~\ref{app:attn-opt}) to avoid materializing the length-$L$ projected key/value tensors (e.g., $XW_K$ and $XW_V$), further reducing memory and length-dependent FLOPs. This substantial FLOPs and memory reduction enables training and serving with ultra-long histories (e.g., 10k-level) within the same compute envelope, thereby improving \emph{scale-up}: at matched computation, STCA can process much longer context and achieves higher accuracy than $O(L^2)$ self-attention that must operate at shorter $L$.

\subsubsection{Input encoding.}
Each historical element $(v_j,a_j)$ is embedded as $\mathbf{x}_j\!\in\!\R^d$ (video, action-type, position fused), and $X=[\mathbf{x}_1,\ldots,\mathbf{x}_L]\!\in\!\R^{L\times d}$. The target video is embedded as $\mathbf{x}_t\!\in\!\R^d$. We use a dimension-preserving \emph{SwiGLUFFN} block (SwiGLU\,+ linear projection), followed by LayerNorm:
\begin{equation}
\label{eq:swigluffn}
\mathrm{SwiGLUFFN}(\mathbf{x})
\;=\;
\Big((\mathbf{x}W_u)\odot\big(\mathbf{x}W_v\odot\mathrm{sigmoid}(\mathbf{x}W_v)\big)\Big)\,W_o,
\end{equation}
where $W_u,W_v\!\in\!\R^{d\times rd}$, $W_o\!\in\!\R^{rd\times d}$, $r\!\ge\!1$, and $\odot$ is element-wise product (biases omitted). We apply the block row-wise to matrices and then normalize:
\begin{align}
\widetilde{X}^{(i)} \;&=\; \mathrm{LN}\!\big(\mathrm{SwiGLUFFN}^{(i)}(X)\big)\ \in\ \R^{L\times d},\\
\mathbf{q}^{(1)} \;&=\; \mathrm{LN}\!\big(\mathrm{SwiGLUFFN}^{(1)}(\mathbf{x}_t)\big)\ \in\ \R^{d},
\end{align}
where $\mathrm{LN}(\cdot)$ denotes LayerNorm.

\paragraph{Multi-head target-to-history cross attention.}
At layer $i$, given $\mathbf{q}^{(i)}$ and $\widetilde{X}^{(i)}$, we compute $h$-head cross attention. Let $d_h{=}d/h$ and
\[
W_Q^{(i,j)},W_K^{(i,j)},W_V^{(i,j)}\in\R^{d\times d_h},\quad W_O^{(i)}\in\R^{d\times d}.
\]
For head $j\in\{1,\ldots,h\}$,
\begin{equation}
\alpha^{(i,j)} \;=\;
\softmax\!\left(\frac{\mathbf{q}^{(i)}W_Q^{(i,j)}\big(\widetilde{X}^{(i)}W_K^{(i,j)}\big)^\top}{\sqrt{d_h}}\right)\in\R^{1\times L},
\end{equation}
\begin{equation}
\mathbf{o}^{(i,j)} \;=\; \alpha^{(i,j)} \big(\widetilde{X}^{(i)}W_V^{(i,j)}\big)\in\R^{1\times d_h}.
\end{equation}
Concatenate heads and project:
\begin{equation}
\label{eq:layer-output}
\mathbf{o}^{(i)} \;=\; \Big[\mathbf{o}^{(i,1)}\parallel\cdots\parallel\mathbf{o}^{(i,h)}\Big]\,W_O^{(i)} \in \R^{d}.
\end{equation}
Per layer, cost is $O(Ldh)$ for a single target query—linear in $L$—versus $O(L^2dh)$ for self-attention over $[t;\mathcal{H}]$.

\subsubsection{Stacking and target-conditioned fusion.}
We \emph{stack} $M$ cross-attention layers and update the query via target-conditioned fusion. To keep dimensions consistent (input $d$), we compress the growing concatenation with a learnable projection:
\begin{equation}
\label{eq:fuse}
\mathbf{q}^{(i+1)} \;=\; \mathrm{LN}\!\left(
\mathrm{SwiGLUFFN}^{(i+1)}\!\Big(\big[\mathbf{o}^{(1)}\parallel\cdots\parallel\mathbf{o}^{(i)}\parallel\mathbf{x}_t\big]\,W_C^{(i+1)}\Big)
\right),
\end{equation}
where $W_C^{(i+1)}\in\R^{(i+1)d\times d}$. 
After $M$ layers, we obtain the layer-wise summaries $\{\mathbf{o}^{(i)}\}_{i=1}^{M}$, which are then combined with the target embedding for prediction.

\subsubsection{Prediction head and objective.}
We form the final target-aware token by compressing all layer summaries with the target:
\begin{equation}
\mathbf{z}\;=\;\mathrm{SwiGLUFFN}\!\Big(\big[\mathbf{o}^{(1)}\parallel\cdots\parallel\mathbf{o}^{(M)}\parallel\mathbf{x}_t\big]\,W_Z\Big),
\quad W_Z\in\R^{(M+1)d\times d}.
\end{equation}
Let
\[
\mathcal{X}_{\mathrm{mix}}
= \mathrm{concat}\!\Big(\mathbf{z},\{\mathbf{u}_k\}_{k=1}^{K},\{\mathbf{c}_\ell\}_{\ell=1}^{C}\Big),
\]
where $\{\mathbf{u}_k\}$ are auxiliary \emph{user-side} tokens (e.g., profile/context features) and $\{\mathbf{c}_\ell\}$ are \emph{candidate-side} tokens associated with the same target $t$ (e.g., content/creator modalities). Then \(\mathrm{RankMixer}\)\cite{DBLP:journals/corr/abs-2507-15551} produces
\begin{equation}
\mathbf{h} = \mathrm{RankMixer}\!\big(\mathcal{X}_{\mathrm{mix}};\Theta\big),\qquad
\hat{y} = \mathrm{sigmoid}\!\big(\mathbf{w}^\top\mathbf{h}+b\big).
\end{equation}
We optimize binary cross-entropy:
\begin{equation}
\label{eq:bce}
\mathcal{L}_{\mathrm{BCE}} = -y\log \hat{y} - (1{-}y)\log (1{-}\hat{y}).
\end{equation}

\subsubsection{Computation optimization for single-query cross attention.}
With exactly one query per layer, we can reorder the computation to avoid
materializing length-$L$ projected key/value tensors (e.g., $XW_K$ and $XW_V$).
Concretely, for $X\!\in\!\R^{L\times d}$ and $q\!\in\!\R^{1\times d}$, the
single-query cross attention can be implemented as
\begin{equation}
\label{eq:attn-optimized}
\mathrm{Attn}(q,X)
\;=\;
\Big(\underbrace{\softmax\!\big(\tfrac{((qW_Q)W_K^\top)X^\top}{\sqrt{d_h}}\big)}_{\alpha\in\R^{1\times L}}\,X\Big)W_V
\;=\;(\alpha X)W_V ,
\end{equation}
which avoids any $L{\times}d_h$ intermediates and reduces length-dependent FLOPs.
The full derivation and FLOPs analysis are provided in Appendix~\ref{app:attn-opt}.

\subsection{Request Level Batching (RLB)}

STCA makes the \emph{per-target} sequence cost linear in history length ($O(L)$), which enables long contexts. However, in real logs each user typically contributes multiple targets within the same request/session. If we still train on independent triplets $(u,t,y)$, the same long history $\mathcal{H}$ is serialized, transferred (CPU$\!\to$GPU), and re-encoded \emph{repeatedly}, so storage and bandwidth—not FLOPs—become the bottleneck as $L$ grows. RLB is therefore the \emph{system-side complement} to STCA: it removes redundant user-side work so that the linear-in-$L$ encoder can scale to multi-k sequences under production budgets.

\subsubsection{Basic idea and unbiasedness.}
RLB aggregates $m$ samples from the same user into a user micro-batch $\mathcal{B}_u=\{(u,t_k,y_k)\}_{k=1}^m$. Let $\Phi_{\text{user}}(\mathcal{H})$ denote the user/history path (shared across targets). With RLB we compute $\Phi_{\text{user}}(\mathcal{H})$ \emph{once} and reuse it for all $\{t_k\}_{k=1}^m$. The per-user loss and overall objective are
\begin{align}
\label{eq:ult-loss}
\mathcal{L}_u &= \frac{1}{m}\sum_{k=1}^{m}\mathcal{L}_{\mathrm{BCE}}\big(\hat{y}(u,t_k),y_k\big),\qquad
\mathcal{L} \;=\; \frac{1}{|\mathcal{U}|}\sum_{u\in\mathcal{U}}\mathcal{L}_u.
\end{align}
\textbf{Unbiasedness.} Writing the conventional objective as a user-average of instance-averages shows that replacing the inner average by a without-replacement average over $m$ targets leaves the expectation unchanged (linearity of expectation). Hence Eq.~\eqref{eq:ult-loss} is an unbiased estimator of empirical risk; RLB changes \emph{only} the computation layout, not the learning objective.
Within-request correlations among targets do not affect this unbiasedness, since RLB only regroups samples and does not change the loss definition.
\subsubsection{Systems view.}
RLB turns repeated user-side work into a \emph{compute-once, reuse-$m$-times} pattern:
we transmit and encode the long history $\mathcal{H}$ once per request and reuse the
shared user/history representation across multiple targets. This reduces redundant
host$\leftrightarrow$device transfer and activation replication, improves kernel
utilization by batching target-side computation, and lowers distributed overhead by
reducing distinct user encodings per step. End-to-end measurements show a
\textbf{77--84\%} bandwidth reduction when enabling RLB (details in
Sec.~\ref{sec:exp-rlt}).

\subsection{Extrapolation: Train Sparsely, Infer Densely}
STCA makes the \emph{per-target} sequence cost linear in $L$ and RLB amortizes the user path across $m$ targets, together enabling long-context \emph{serving} under strict latency and bandwidth budgets. However, \emph{training} on uniformly long histories still scales linearly with the number of tokens processed per batch, quickly exhausting memory and throughput as $L$ grows. To decouple training cost from deployment-time context length, we introduce a length–extrapolation regimen that \emph{trains sparsely} (low average tokens per batch) yet \emph{infers densely} (long histories at test time).
Throughout this section we fix the deployment target at 
$L_{\text{infer}}= \mathbf{10\mathrm{k}}$ and the training average at 
$L_{\text{train}}^{\text{avg}}=\mathbf{2\mathrm{k}}$, yielding an extrapolation ratio 
$\rho_{\text{extra}}=\frac{L_{\text{infer}}}{L_{\text{train}}^{\text{avg}}}=\mathbf{5}$.

We therefore follow the \emph{Stochastic Length} (SL) training paradigm during training, each input sequence is randomly truncated to a length \(L_{\text{train}} \in [L_{\text{train}}^{\min}, L_{\text{train}}^{\max}]\), where \(L_{\text{train}}^{\max} \leq L_{\text{infer}}\), and \(L_{\text{infer}}\) is the maximum sequence length used at inference. We quantify the resulting efficiency via \emph{sequence sparsity} (SS), defined as  
\begin{equation}
    \text{SS} = \frac{\mathbb{E}[L_{\text{train}}]}{L_{\text{train}}^{\max}} = \frac{L_{\text{train}}^{\text{avg}}}{L_{\text{train}}^{\max}},
\end{equation}
which reflects the average computational cost relative to the maximum training length.

Under the Stack Cross-Attention (STCA) architecture, this stochastic training strategy raises two key challenges:  
\begin{enumerate}
    \item \textbf{Batch-Level Load Balancing}: Variable-length sequences cause GPU workload imbalance, as batch processing time is dictated by the longest sequence—undermining potential FLOPs savings.
    \item \textbf{Subsequence Selection}: An effective selection strategy must minimize training sequence length (i.e., minimize SS) without degrading model accuracy.  
\end{enumerate}

\subsubsection{Subsequence Selection Strategy}
\label{sec:subsequence-selection}

The subsequence selection process involves two steps: (1) sampling a stochastic training length \(L_{\text{train}}\), and (2) selecting which elements from the full history to retain.

\paragraph{Stochastic Length Sampling.}
We sample a stochastic training length $L_{\text{train}}\!\in\![L_{\text{train}}^{\min},\,L_{\text{train}}^{\max}]$
using a \emph{U-shaped} Beta distribution over the normalized ratio, so that the curriculum
mixes very short windows with occasional long windows while keeping the average length
at $L_{\text{train}}^{\text{avg}}$ (thereby controlling $\text{SS}$ and $\rho_{\text{extra}}$).
This length sampling acts as a data curriculum; the end-to-end objective remains the same
BCE loss in Eq.~\eqref{eq:bce}. We defer the exact sampling formula, expectation constraint,
and hardware-alignment details to Appendix~\ref{app:extrapolation-details}.

\paragraph{Element Selection Policy.}  
Given \(L_{\text{train}}\), we select the corresponding number of items from the user's full history (truncated to \(L_{\text{infer}}\) at inference). \emph{Empirical results show that retaining the most recent \(L_{\text{train}}\) interactions—i.e., the temporal suffix—consistently yields optimal accuracy.}

\subsubsection{Batch-Level Load Balancing}
Variable-length $L_{\text{train}}$ can cause batch-level workload imbalance (step time dominated by the longest
sequence). We therefore apply a batch-level load-balancing operator that keeps the total token budget close to
$B\cdot L_{\text{train}}^{\text{avg}}$ (batch size $B$), and compacts sequences to reduce padding while preserving
the stochastic-length curriculum. We implement the resulting ragged computation with an index-based target-attention
kernel to avoid padding overhead. Full operator details and the ragged attention implementation are provided in
Appendix~\ref{app:extrapolation-details}.

\section{Experiments}
\subsection{Offline Comparison on Douyin}
\label{sec:exp-douyinlite}

\subsubsection{Setup.}
We evaluate sequence encoders on the \emph{Douyin} offline dataset for three objectives—\textbf{finish} (completion), \textbf{skip} (quick skip), and \textbf{head} (creator-page click)—reporting \emph{AUC} (higher is better) and \emph{NLL} (lower is better). To keep the comparison conservative, \emph{all baselines are augmented with TWIN(10k)\cite{DBLP:conf/cikm/SiGSZLHCYZLZZN024}} (a retrieval-style block built from a 10k-length behavior search), whereas \emph{our method removes TWIN(10k)} and relies purely on end-to-end long-history modeling. This setup is intentionally conservative and \emph{favors the baselines}, since they receive extra retrieval signals.
Our claim is that STCA+RLB+Ext can \emph{replace} such a heavy retrieval block at comparable end-to-end cost while preserving full differentiability over long histories.
\footnote{All models share identical non-sequence features, optimizers, and data splits; only the sequence encoder and the use of TWIN(10k) differ.}
Importantly, we compare under \emph{roughly matched compute}: per-sample sequence FLOPs and step-time are aligned across methods. For quadratic-cost encoders (Transformer, HSTU), we \emph{reduce depth/width} to keep their total compute comparable to STCA, ensuring fair comparisons.

We compare \emph{Single-layer target attention}, \emph{DIN}, \emph{Transformer} (self-attention), and \emph{HSTU} against our \textbf{STCA} (stacked target$\!\to$history cross attention) with \textbf{RLB} and \textbf{Train Sparsely / Infer Densely} (``Ext''). Table cells report \emph{\% changes} vs.\ a production baseline: \emph{RankMixer + Single-layer Target Attention + TWIN(10k)}; positive $\Delta$AUC and negative $\Delta$NLL are better.

\begin{table}[t]
\centering
\caption{Offline results on \textbf{Douyin}. All values are in \%.}
\label{tab:douyinlite-main}
\setlength{\tabcolsep}{6pt}
\resizebox{\linewidth}{!}{%
\begin{tabular}{lrrrrrr}
\toprule
\multirow{2}{*}{Model} & \multicolumn{2}{c}{\textbf{Finish}} & \multicolumn{2}{c}{\textbf{Skip}} & \multicolumn{2}{c}{\textbf{Head}} \\
\cmidrule(lr){2-3}\cmidrule(lr){4-5}\cmidrule(lr){6-7}
& $\Delta$AUC$\uparrow$ & $\Delta$NLL$\downarrow$ & $\Delta$AUC$\uparrow$ & $\Delta$NLL$\downarrow$ & $\Delta$AUC$\uparrow$ & $\Delta$NLL$\downarrow$ \\
\midrule
Baseline & $0.00$ & $0.00$ & $0.00$ & $0.00$ & $0.00$ & $0.00$ \\
DIN                   & $+0.19$ & $-0.17$ & $+0.23$ & $-0.18$ & $+0.19$ & $-0.21$ \\
Trans                 & $+0.25$ & $-0.46$ & $+0.27$ & $-0.36$ & $+0.38$ & $-0.27$ \\
HSTU                  & $+0.31$ & $-0.86$ & $+0.52$ & $-0.62$ & $+0.36$ & $-0.43$ \\
\textbf{Ours}         & $\mathbf{+0.49}$ & $\mathbf{-1.16}$ & $\mathbf{+0.71}$ & $\mathbf{-1.14}$ & $\mathbf{+0.39}$ & $\mathbf{-1.41}$ \\
\bottomrule
\end{tabular}
}
\vspace{2pt}
\end{table}

\subsubsection{Results and discussion.}
In Table \ref{tab:douyinlite-main}, \textbf{STCA+RLB+Ext} achieves the strongest improvements on all tasks despite using \emph{no} TWIN(10k): \textbf{+0.49}/\textbf{-1.16} (finish), \textbf{+0.71}/\textbf{-1.14} (skip), and \textbf{+0.39}/\textbf{-1.41} (head). Baselines also improve under the matched-compute setting—DIN up to \textbf{+0.23}/\textbf{-0.21}, Transformer up to \textbf{+0.38}/\textbf{-0.46}, HSTU up to \textbf{+0.52}/\textbf{-0.86}—yet our method consistently delivers the largest lifts, especially in NLL (e.g., head \textbf{-1.41}). 

The gains align with our design: retrieval features pre-select and lose information and end-to-end gradients; \textbf{STCA} performs \emph{exact} softmax attention over the full history with $O(L)$ cost per target, \textbf{RLB} removes redundant user encodings across targets, and \textbf{Train Sparsely / Infer Densely} exposes a calibrated tail of long contexts to enable multi-thousand-token inference without full-length training.

\paragraph{Takeaway.}
Under identical non-sequence features and \emph{matched compute}, \textbf{STCA+RLB+Ext} improves both ranking and calibration while simplifying the stack (no TWIN(10k)), providing a practical path to accurate, deployable long-sequence modeling on Douyin.

\subsection{Ablation: Accuracy vs.\ Model Complexity}
\label{sec:exp-stca}

\begin{table*}[!t]
\centering
\caption{Ablation of the 4L complex STCA at 512 tokens.}
\label{tab:stca-ablation}
\begin{tabular}{@{}lcl@{}}
\toprule
Component & $\Delta$ AUC & Notes \\
\midrule
Enlarge sparse ID embedding: $128\!\rightarrow\!320$ & $+0.08\%$ &
Video-ID embedding width. Use smaller init range and LR to avoid instability. \\
Add sequence-side FFN; depth $2\!\rightarrow\!4$ & $+0.18\%$ &
Token-wise FFN on the history path; then double STCA depth from 2L to 4L. \\
FFN $\rightarrow$ SwiGLU & $+0.11\%$ &
Upgrade FFNs to SwiGLU; widen hidden by $\times 2$. \\
Attention heads: $8\!\rightarrow\!16$ & $+0.05\%$ &
More heads improve target-conditioned selectivity with modest cost. \\
Query fusion & $+0.06\%$ &
At layer $i{+}1$, concatenate $[\mathbf{o}^{(1)}\ldots \mathbf{o}^{(i)},\mathbf{x}_t]$ then apply SwiGLU (Eq.~\ref{eq:fuse}). \\
Time-delta side info & $+0.08\%$ &
Add per-token feature: request time minus item timestamp (recency prior).  \\
\bottomrule
\end{tabular}
\end{table*}

\paragraph{Setup and metrics.}
We evaluate the end-to-end stack \textbf{STCA} $\rightarrow$ \textbf{RankMixer} with \textbf{RLB} ($m{=}8$) and single-query cross-attention (Sec.~\ref{sec:method}). Unless noted, training and evaluation use $L{=}512$; we report \emph{finish AUC} lift (\%) over a strong \textbf{RankMixer}-only baseline trained with identical non-sequence features and optimization.

\paragraph{Where the gains come from.}
Ablations in Table~\ref{tab:stca-ablation} indicate that adding a token-wise FFN on the sequence path and deepening STCA from 2L$\!\to\!$4L provides the largest single boost ($+0.18\%$). Upgrading FFNs to \emph{SwiGLU} yields a further $+0.11\%$. Enlarging sparse ID embeddings ($+0.08\%$) and introducing time-delta side information ($+0.08\%$) both contribute meaningfully. Increasing attention heads from $8\!\to\!16$ offers a modest but positive gain ($+0.05\%$). Finally, the \emph{query fusion} mechanism (Eq.~\ref{eq:fuse}) adds $+0.06\%$ by reinjecting lower-layer summaries into higher layers for target-conditioned reasoning.

\begin{figure}[t]
  \centering
  \includegraphics[width=\linewidth]{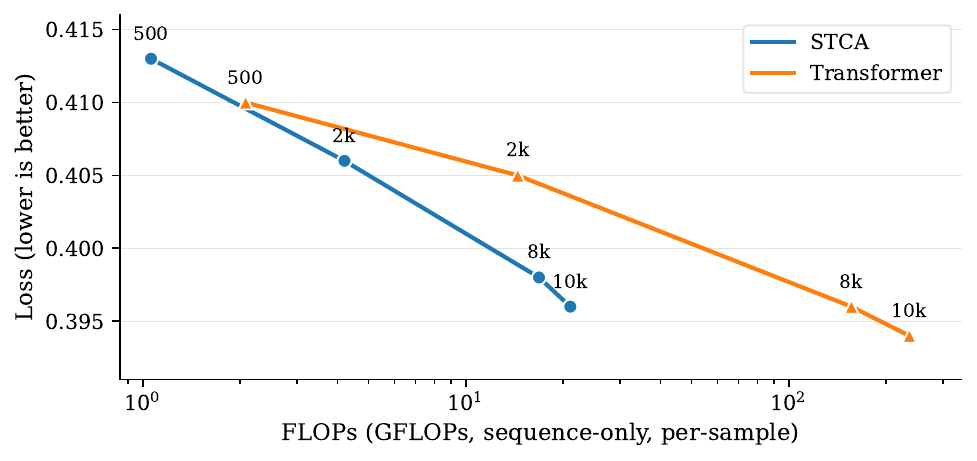}
  \caption{Compute--quality scaling of \textbf{STCA} vs \textbf{Transformer}.
  X-axis: per-sample \emph{sequence-only} forward FLOPs (log scale). Y-axis: NLL (lower is better). Markers show $L\!\in\!\{500,2\mathrm{k},8\mathrm{k},10\mathrm{k}\}$. Both use 4 layers with $d{=}256$, $h{=}8$, $r{=}4$.}
  \label{fig:flops-vs-loss}
\end{figure}

\paragraph{Compute--quality scaling.}
Figure~\ref{fig:flops-vs-loss} couples compute (FLOPs) with quality (NLL) under matched depth/width. Two points stand out:
\begin{itemize}[leftmargin=1.15em,itemsep=2pt]
  \item \textbf{Linear vs.\ quadratic in $L$.} STCA’s sequence-side FLOPs grow linearly; Transformer’s grow quadratically. From $L{=}500\!\to\!10$k, STCA rises $1.06\!\to\!21.06$ GFLOPs ($\sim\!19.9{\times}$), while Transformer rises $2.08\!\to\!236.26$ GFLOPs ($\sim\!113.6{\times}$).
  \item \textbf{Better frontier at long $L$.} At similar NLL ($\approx\!0.396$), STCA runs at $L{=}10$k with $21.06$ GFLOPs, whereas Transformer needs $L{=}8$k and $156.24$ GFLOPs ($\sim\!7.4{\times}$ higher).
\end{itemize}
These results align with our broader findings: \textbf{STCA} removes history self-attention and focuses compute on target$\leftrightarrow$history relevance (near-linear scaling), \textbf{RLB} amortizes the user path (and extends across requests/sessions), and \textbf{train sparsely / infer densely} trains at $\sim\!2$k but serves up to $10$k. Combined, they provide a practical scale-up path: STCA sustains longer contexts at similar or lower FLOPs while preserving or improving ranking quality.

\subsection{Request-Level Batching (RLB)}
\label{sec:exp-rlt}

\paragraph{Setup.}
RLB realizes user-centric batching at the \emph{request} boundary: for each request containing one user history and multiple targets, we transmit and encode the \emph{shared} user/context payload once, reuse it for all targets, and aggregate gradients at the request level before synchronization. Unless otherwise noted, the stack is STCA\,$\rightarrow$\,RankMixer with RLB enabled (Sec.~\ref{sec:method}), trained on the same data, optimizer, and batch size as the point-wise baseline. Reported bandwidth figures include \emph{all} non-sequence features (profile/context, content, creator, etc.), i.e., end-to-end payloads.

\paragraph{Bandwidth footprint.}
RLB substantially cuts inter-module traffic by amortizing the $O(L)$ user/history payload across targets: the measured reduction is \textbf{77\%} at history length $L{=}512$ and \textbf{84\%} at $L{=}2\mathrm{k}$ (both including all existing features). The larger saving at longer $L$ follows directly from avoiding per-target retransmission of the same history.

\paragraph{Throughput and scalability.}
Relative to a point-wise baseline (\(1\times\)), RLB delivers a \textbf{2.2\(\times\)} end-to-end training throughput gain. With further kernel optimizations—specialized batched matmul for the reordered single-query attention, high-throughput \emph{SwiGLU}, and optimized LayerNorm—the speedup reaches \textbf{5.1\(\times\)}. Under the same infrastructure, amortizing per-target activations and payload also raises the \emph{maximum trainable sequence length} by roughly \textbf{8\(\times\)} (workloads that were memory/IO bound at length \(L\) become trainable at \(\approx 8L\)).

\paragraph{Parameter server and inter-module costs.}
At synchronization and feature-serving boundaries, RLB further lowers overhead: we observe a \textbf{50\%} reduction in Parameter Server (PS) CPU usage during training and a \textbf{50\%} reduction in data$\leftrightarrow$training communication bandwidth; training-side PS CPU load likewise drops by \textbf{50\%}.

\paragraph{Discussion.}
These gains complement STCA’s architectural efficiency. STCA removes the quadratic dependence on history length via single-query cross attention ($O(L)$), while RLB eliminates \emph{redundant} movement and re-encoding of the same user/history across targets. Together they yield higher samples-per-second, lower PS contention, and much larger feasible sequence lengths, enabling long-context models within fixed latency and hardware budgets.

\subsection{Extrapolation: Train Sparsely, Infer Densely}
\label{sec:exp-extrapolation}

\paragraph{Experimental Setup.}
We evaluate the proposed extrapolation framework using STCA encoder with single-query optimization. The user token $\mathbf{z}$ is fed into \emph{RankMixer} (Sec.~\ref{sec:method}) with RLB training ($m{=}8$). During training, history length $L_{\text{train}}$ is randomized following Sec.~\ref{sec:subsequence-selection}. All results report finish AUC improvements versus a fixed $2\mathrm{k}$-token baseline. Unless otherwise specified, \textbf{inference uses $L_{\text{infer}}=\mathbf{10\mathrm{k}}$}.

\begin{table}[t]
\centering
\caption{Effect of maximum training length.}
\label{tab:extrapolation-lmax}
\begin{tabular}{@{}lccc@{}}
\toprule
Metric & $L_{\text{train}}^{\max}=2\mathrm{k}$ & $L_{\text{train}}^{\max}=4\mathrm{k}$ & $L_{\text{train}}^{\max}=10\mathrm{k}$ \\
\midrule
Finish AUC lift & $+0.03\%$ & $+0.09\%$ & $+0.21\%$ \\
\bottomrule
\end{tabular}
\end{table}

\paragraph{The impact of the maximum length (Table \ref{tab:extrapolation-lmax}).} Progressively increasing $L_{\text{train}}^{\max}$ from $2\mathrm{k}$ to $10\mathrm{k}$ yields a corresponding improvement in the AUC lift, which rises from a marginal $+0.03\%$ to a more substantial $+0.21\%$. These results confirm that to ensure robust generalization to long contexts, the training regime must expose the model to sequence lengths that closely match or approximate those encountered during inference.

\begin{table}[t]
\centering
\caption{Effect of average training length.}
\label{tab:extrapolation-mean}
\begin{tabular}{@{}lccc@{}}
\toprule
Metric & $L_{\text{train}}^{\text{avg}}=1.0\mathrm{k}$ & $L_{\text{train}}^{\text{avg}}=2.0\mathrm{k}$ & $L_{\text{train}}^{\text{avg}}=2.5\mathrm{k}$ \\
\midrule
Finish AUC lift & $+0.09\%$ & $+0.21\%$ & $+0.22\%$ \\
Sequence Sparsity & 10.0\% & 20.0\% & 25.0\% \\
\bottomrule
\end{tabular}
\end{table}

\paragraph{Sequence Sparsity Optimization.}
Table~\ref{tab:extrapolation-mean} shows the efficiency-accuracy trade-off. Increasing $L_{\text{train}}^{\text{avg}}$ from $1.0\mathrm{k}$ to $2.5\mathrm{k}$ improves AUC lift from $+0.09\%$ to $+0.22\%$ while increasing sequence sparsity (SS) from 10\% to 25\%. The diminishing returns beyond $\bar{L}=2.0\mathrm{k}$ confirm that $\text{SS} \approx 20\%$ provides the optimal balance. Compared to SL in HSTU (SS=57.6\%)~\cite{DBLP:conf/icml/ZhaiLLWLCGGGHLS24}, we achieve better computational efficiency.

\paragraph{Subsequence Selection Strategy Validation.} Retaining the most recent interactions (greedy) achieves $+0.21\%$ AUC improvement, while random sampling yields no gain, strongly supporting the importance of temporal locality.

\begin{table}[t]
\centering
\caption{Beta distribution shape parameter analysis.}
\label{tab:beta-shape}
\begin{tabular}{@{}lccc@{}}
\toprule
Metric & $\alpha=0.02$ &  $\alpha=0.5$ &  $\alpha=10$ \\
\midrule
Finish AUC lift & $+0.21\%$ & $+0.11\%$ & $+0.08\%$ \\
Shape & U-shaped  & Decreasing & Skewed \\
\bottomrule
\end{tabular}
\end{table}

\paragraph{Beta Distribution Shape Analysis.}
Table~\ref{tab:beta-shape} validates our distribution design: the U-shaped distribution ($\alpha=0.02$) achieves superior performance ($+0.21\%$) compared to decreasing ($+0.11\%$) and skewed ($+0.08\%$) distributions, confirming that bimodal sampling optimizes the training curriculum.

\paragraph{Efficiency-Accuracy Trade-off.}
Our approach achieves $+0.23\%$ offline AUC improvement at $10\mathrm{k}$ inference, capturing $\sim80\%$ of the full $10\mathrm{k}$ training gain ($+0.30\%$) at one-third computational cost. Online A/B tests confirm production viability ($+0.17\%$ finish AUC).

\paragraph{Discussion.}
The results comprehensively validate our methodology: (1) Stochastic training with U-shaped Beta distribution enables effective extrapolation; (2) Temporal suffix selection preserves sequential patterns; (3) Load-balancing ensures training efficiency. Under our default setting of $L_{\text{train}}^{\text{avg}}=\mathbf{2\mathrm{k}}$, 
$L_{\text{train}}^{\max}=\mathbf{10\mathrm{k}}$, and $L_{\text{infer}}=\mathbf{10\mathrm{k}}$ ($\rho_{\text{extra}}=\mathbf{5}$). Compared to HSTU's SL (SS=57.6\%), we reduce SS to 20\% while maintaining accuracy, demonstrating superior computational efficiency.

\begin{table*}[t]
\centering
\caption{One-month online A/B lifts (\%) over control on \textbf{Douyin} and \textbf{Douyin Lite}.}
\label{tab:online-ab}
\begin{tabular}{@{}lccccc ccccc@{}}
\toprule
& \multicolumn{5}{c}{\textbf{Douyin}} & \multicolumn{5}{c}{\textbf{Douyin Lite}} \\
\cmidrule(lr){2-6}\cmidrule(lr){7-11}
\textbf{Segment} & \textbf{30-day Act.} & \textbf{Stay Time} & \textbf{Finish} & \textbf{Comment} & \textbf{Like} & \textbf{30-day Act.} & \textbf{Stay Time} & \textbf{Finish} & \textbf{Comment} & \textbf{Like} \\
\midrule
Low   & 0.3659\% & 2.0070\% & 5.4987\% & 2.6848\% & 2.4367\% & 0.3575\% & 1.3888\% & 6.2808\% & 6.3385\% & 3.1623\% \\
Medium& 0.3788\% & 1.7065\% & 5.2062\% & 0.7028\% & 2.3779\% & 0.2832\% & 1.3172\% & 5.9688\% & 5.7934\% & 5.6372\% \\
High  & 0.1396\% & 1.1262\% & 3.7973\% & 1.4012\% & 2.1455\% & 0.1604\% & 0.9872\% & 4.9922\% & 2.8496\% & 3.1752\% \\
\midrule
\textbf{All Users} & \textbf{0.1161\%} & \textbf{0.9266\%} & \textbf{3.3454\%} & \textbf{1.5678\%} & \textbf{1.8282\%} & \textbf{0.1281\%} & \textbf{0.8467\%} & \textbf{4.2275\%} & \textbf{2.6167\%} & \textbf{2.3828\%} \\
\bottomrule
\end{tabular}%

\end{table*}

\subsection{Online A/B Results}
\label{sec:exp-online}

\paragraph{Setup.}
We deployed \textbf{STCA + RLB + Extrapolation} (Sec.~\ref{sec:method}) for one month on \emph{Douyin} and \emph{Douyin Lite}, replacing TWIN(10k)-augmented retrieval features with our single-query target$\!\to$history encoder while keeping all other components unchanged. We report percentage lifts over control on \textbf{30-day Activeness}, \textbf{App Stay Time}, \textbf{Finish}, \textbf{Comment}, and \textbf{Like}, overall and by user-activity segment (Table \ref{tab:online-ab}).

\paragraph{Inference cost.} In our internal accounting, removing TWIN increases GPU cost by $+33\%$ but reduces CPU cost by $-16\%$, leading to an estimated net total cost change of about $+17\%$, which we consider acceptable given the online gains.

\paragraph{Findings.}
Our method delivers \emph{consistent and sizable} online gains across both products and all segments. Overall, \textbf{Finish} and \textbf{App Stay Time} improve together, and interactive signals (\textbf{Comment}, \textbf{Like}) rise markedly. Gains are strongest for \emph{low/medium-activity} users, indicating better personalization under sparse/noisy recent behavior, while \textbf{30-day Activeness} also increases modestly yet consistently.
The improvements stem from (i) \textbf{STCA} focusing compute on exact target$\!\to$history interactions over long contexts, (ii) \textbf{RLB} amortizing user encoding to keep serving costs within budget, and (iii) \textbf{Train Sparsely/Infer Densely} exposing a calibrated tail of long sequences during training. Together, these make end-to-end long-history modeling both accurate and deployable at scale.

\section{Related Work and Discussion}\label{sec:related}

\subsection{Modeling User Behavior Sequences}
Early systems modeled sequences with non-deep methods such as item-to-item CF and Markov transitions~\cite{DBLP:journals/internet/LindenSY03,DBLP:conf/www/RendleFS10}. Deep learning then brought session-based RNNs and industrial two-stage stacks (e.g., YouTube)~\cite{DBLP:journals/corr/HidasiKBT15,DBLP:conf/recsys/CovingtonAS16}. Attention/Transformer models became dominant: DIN/DIEN condition history on the candidate to emphasize target-aware selection~\cite{DBLP:conf/kdd/ZhouZSFZMYJLG18,DBLP:conf/aaai/ZhouMFPBZZG19}, while SASRec/BERT4Rec provide strong self-attention baselines that model temporal dependencies through bidirectional or unidirectional attention~\cite{DBLP:conf/icdm/KangM18,DBLP:conf/cikm/SunLWPLOJ19}; BST further demonstrated online deployment in an industrial feed-ranking setting~\cite{DBLP:journals/corr/abs-1905-06874}. In parallel, multi-interest modeling explicitly captures diverse user intents and has become widely used in large-scale platforms~\cite{DBLP:conf/cikm/LiLWXZHKCLL19}. 

Recent production work emphasizes \emph{long} histories and system compatibility: user-representation approaches (e.g., PinnerFormer) compress long-term behaviors into condensed memories, and real-time/batch fusion (e.g., TransAct) merges streaming signals with precomputed features for latency-aware serving~\cite{DBLP:conf/kdd/PanchaZLR22,DBLP:conf/kdd/XiaEPBWGJFZZ23,DBLP:journals/corr/abs-2506-02267}. To scale further, two-stage paradigms retrieve a target-relevant slice before fine modeling (SIM, UBR4CTR)~\cite{DBLP:conf/cikm/PiZZWRFZG20,DBLP:conf/sigir/Qin0WJF020}, or engineer sparse memories/hierarchies (SAMN, TWIN/TWIN-V2) to bound the cost of processing very long logs~\cite{DBLP:conf/cikm/LinZWDCW22,DBLP:conf/kdd/ChangZFZGLHLNSG23,DBLP:conf/cikm/SiGSZLHCYZLZZN024}. These designs typically trade exact end-to-end gradients over the full sequence for efficiency via truncation, retrieval, or summarization.
\emph{Compared with these, our STCA performs single-query target$\to$history cross attention end-to-end over the full sequence, achieving linear complexity in $L$ and avoiding retrieval/truncation.} This yields a practical path to \emph{very} long histories in production while maintaining differentiability through all observed interactions.

\subsection{Organizing Training Samples}
\label{sec:rw-training-organization}
Classic training formulations include pointwise/pairwise learning (e.g., BPR)~\cite{DBLP:conf/uai/RendleFGS09}, large-batch CTR pipelines (DLRM, production systems)~\cite{DBLP:journals/corr/abs-1906-00091,DBLP:journals/corr/abs-2003-09518}, and session-parallel batching for sequences~\cite{DBLP:journals/corr/HidasiKBT15}. At industrial scale, training is often I/O/memory bound rather than FLOP bound~\cite{DBLP:conf/isca/MudigereHHJT0LO22}, motivating truncation or retrieval-first construction~\cite{DBLP:conf/cikm/PiZZWRFZG20,DBLP:conf/sigir/Qin0WJF020} and engineering for ultra-long contexts~\cite{DBLP:conf/cikm/LinZWDCW22,DBLP:conf/kdd/ChangZFZGLHLNSG23,DBLP:conf/cikm/SiGSZLHCYZLZZN024}. Modern stacks (e.g., TorchRec) provide jagged tensors and sharding to better utilize hardware and manage sparsity~\cite{torchrec22}, and production systems mix real-time/batch signals to balance freshness and stability~\cite{DBLP:conf/kdd/XiaEPBWGJFZZ23,DBLP:conf/kdd/PanchaZLR22}. \emph{Against this backdrop, our Request Level Batching (RLB)} reorganizes data at the user/request granularity: compute the user/history encoder once per request and reuse it across $m$ targets, cutting the per-target user-path cost from $O(L)$ to $\approx O(L/m)$ while keeping the empirical-risk objective unbiased. Unlike truncation/retrieval, RLB preserves the full history and amortizes movement and recomputation, directly addressing the dominant I/O and activation bottlenecks at scale~\cite{DBLP:conf/isca/MudigereHHJT0LO22}. It is orthogonal to retrieval, kernel-level speedups, and sharding (e.g., jagged tensors in~\cite{torchrec22}) and aligns with request-level grouping in production, improving bandwidth, peak memory, and kernel efficiency without changing the loss.

\subsection{Length Extrapolation}
\label{sec:rw-length-extrapolation}
LLMs show that \emph{``train sparsely, infer densely''} is feasible via positional designs (ALiBi, RoPE) and memory/attention patterns (Transformer - XL, Longformer, BigBird) that enable generalization beyond the training window~\cite{DBLP:conf/iclr/PressSL22,DBLP:journals/ijon/SuALPBL24,DBLP:conf/acl/DaiYYCLS19,DBLP:journals/corr/abs-2004-05150,DBLP:conf/nips/ZaheerGDAAOPRWY20}. \emph{Our approach adapts this spirit to recommendation} by (i) replacing quadratic history self-attention with exact single-query cross attention (STCA), and (ii) reorganizing training via RLB to amortize user encoding. Unlike retrieval pipelines~\cite{DBLP:conf/cikm/PiZZWRFZG20,DBLP:conf/sigir/Qin0WJF020,DBLP:conf/kdd/ChangZFZGLHLNSG23,DBLP:conf/cikm/SiGSZLHCYZLZZN024}, we keep end-to-end gradients over the full history; unlike kernel/sparsity methods~\cite{DBLP:journals/corr/abs-2004-05150,DBLP:conf/nips/ZaheerGDAAOPRWY20}, we directly match the target-over-history interaction while aligning with production constraints~\cite{DBLP:conf/kdd/PanchaZLR22,DBLP:conf/kdd/XiaEPBWGJFZZ23,DBLP:conf/isca/MudigereHHJT0LO22,torchrec22}. 

Concretely, we use stochastic-length sampling during training to keep the \emph{average} sequence length short while exposing a calibrated tail of longer contexts, and we serve at much longer histories at inference. This regimen leverages STCA’s linear-in-$L$ compute and RLB’s amortization to make extrapolation practical in production, without truncation or surrogate memories.

\section{Conclusion}
We present an end-to-end recipe for long-sequence recommendation that is simultaneously architectural, system, and training efficient. Architecturally, STCA replaces history self-attention with single-query target$\to$history cross-attention, yielding linear ($O(L)$) complexity in sequence length. System-wise, RLB reuses the user-side encoding at the request level, eliminating redundant transfers and computation. Training-wise, a ``train sparsely / infer densely'' regimen enables dense inference on long histories at modest training cost. Offline and online experiments show stable, substantial gains and scaling-law–like improvements as sequence length and sequence-module capacity grow. Key findings include: single-query attention is sufficient for short-video ranking while preserving $O(L)$ cost; stochastic-length sampling with a small-$\alpha$ Beta achieves roughly 80\% of the 10k-window benefit at about one-third the training cost; parameter budget is best spent on the sequence path via SwiGLU, cross-layer query fusion, and time-delta features—whose impact increases with longer contexts; and system levers are decisive for deployability—at ($L{=}2\mathrm{k}$), RLB reduces end-to-end bandwidth by up to ~84\%, halves PS CPU usage, delivers ~2.2$\times$ throughput, and expands the maximum trainable sequence length by ~8$\times$. In production, we observe a $1.6\%$ increase in the average number of clustered content categories per user, suggesting slightly higher diversity.

In practice, a strong operating point couples a 4-layer STCA (with SwiGLU and query fusion) with time-delta features, RLB, and stochastic-length training (small $\alpha$, \emph{training mean} $\approx 2\mathrm{k}$) and \emph{inference length} $10\mathrm{k}$ (a $5\times$ extrapolation ratio). This configuration provides monotonic improvements with both sequence length and sequence-module capacity while keeping training and serving within budget, offering a production-ready path to accurate long-sequence recommendation.

\bibliographystyle{ACM-Reference-Format}
\bibliography{bib}

\appendix

\section{Additional Implementation and System Details}
\label{app:impl-details}

This appendix provides additional implementation and system details that are omitted from the main text for brevity.
We organize the material as follows: (i) training pipeline and modular integration, (ii) STCA implementation details
and attention efficiency, (iii) request-level batching (RLB) as a sample-organization primitive, and (iv) the full
extrapolation procedure, including Beta sampling and batch-level load balancing.

\subsection{Training Pipeline and Modular Integration}
\label{app:training-pipeline}
Directly training at long contexts can be unstable (e.g., $L{=}2048$). We therefore adopt a simple curriculum:
(i) pretrain at $L{=}512$ to establish robust token-level filters and attention patterns; (ii) continue training at
$L{=}2048$. During architecture iteration, we prototype at $L{=}512$ for faster convergence and lower resource use.
For integration into a larger production stack, we first train the sequence sub-network to convergence, load its
parameters into the composite model, and finally perform joint finetuning. This staging mitigates vanishing gradients
along the sequence path when the rest of the stack is already strong.

\subsection{STCA Details: Robustness and Attention Efficiency}
\label{app:stca-details}

\subsubsection{Generalization and robustness of STCA}
STCA generalizes well across sequence lengths because each history token is processed independently \emph{conditioned
on the target}, making the architecture naturally length-agnostic; stacking enables information aggregation to scale
smoothly as $L$ increases. By filtering history through the target at every layer, STCA is less sensitive to irrelevant
or noisy behaviors common in real logs. This improves robustness to variable-length sequences and heterogeneous user
patterns, which is crucial for industrial deployment at scale.

\subsubsection{Single-query attention reordering and FLOPs analysis}
\label{app:attn-opt}
With exactly one query per layer, let $X\!\in\!\R^{L\times d}$, $q\!\in\!\R^{1\times d}$, and $d_h{=}d/h$.
The standard cross-attention form is
\begin{equation}
\label{eq:attn-standard}
\mathrm{Attn}(q,X)
\;=\;
\softmax\!\Big(\tfrac{(qW_Q)(XW_K)^\top}{\sqrt{d_h}}\Big)\cdot(XW_V),
\end{equation}
which projects all $L$ tokens twice and materializes the length-$L$ tensors $XW_K$ and $XW_V$.

\paragraph{Reordering.}
We can reorder the computation to remove the length-$L$ projections:
\[
u \,=\, (qW_Q)W_K^\top \in \R^{1\times d},\quad
\alpha \,=\, \softmax\!\Big(\tfrac{u\,X^\top}{\sqrt{d_h}}\Big)\in\R^{1\times L},
\]
and then compute
\[
o \,=\, (\alpha X)\,W_V \in \R^{1\times d_h},
\quad W_Q,W_K,W_V\!\in\!\R^{d\times d_h}.
\]
This yields the equivalent implementation summarized in Eq.~\eqref{eq:attn-optimized}:
\begin{equation*}
\mathrm{Attn}(q,X)
\;=\;
\Big(\softmax\!\big(\tfrac{((qW_Q)W_K^\top)X^\top}{\sqrt{d_h}}\big)\,X\Big)W_V
\;=\;(\alpha X)W_V .
\end{equation*}

\paragraph{FLOPs and intermediates.}
Per head, the reordered path costs $O(d\,d_h)+O(Ld)+O(d\,d_h)$; across $h$ heads it totals $O(Ldh + d^2)$ while
avoiding any $L{\times}d_h$ intermediates. In contrast, the naïve path spends $\approx 4Ldd_h$ FLOPs on forming
$(XW_K,\,XW_V)$ and materializes two $L{\times}d_h$ tensors per head. The reordered path replaces them with a single
weighted reduction $\alpha X$ of cost $2Ld$ FLOPs and no $L{\times}d_h$ intermediates. Thus the length-dependent FLOPs
shrink by a factor of $\approx 2d_h{=}\,2d/h$.

\paragraph{Example.}
With $d{=}256$ and $h{=}8$ ($d_h{=}32$), the length-dependent FLOPs reduction is $\sim\!2d_h=\sim\!64\times$.

\subsubsection{Relation to attention optimizers and fused kernels}
GQA/MQA reduce the number of distinct K/V projections to save memory/bandwidth, but their scoring over a length-$L$
sequence remains $O(L^2 d)$. IO-efficient kernels (e.g., FlashAttention) reduce memory traffic yet keep quadratic
compute; linear/low-rank variants reach $O(Ld)$ via approximations. \textbf{STCA} instead removes history
self-interactions entirely and performs \emph{exact} single-query target$\to$history attention with $O(Ldh)$ per layer
(further benefiting from the reordering in Eq.~\eqref{eq:attn-optimized}), while the downstream RankMixer operates on a
small, length-independent token set. \textbf{RLB} complements this by amortizing the user path across $m$ targets,
effectively turning $O(L)$ into $O(L/m)$ per target. These techniques are compatible with head sharing and fused kernels,
but our dominant savings arise from (i) architectural removal of the quadratic term and (ii) system-level amortization,
achieved \emph{without} retrieval or truncation.

\subsection{Request-Level Batching (RLB) as Sample Organization}
\label{app:rlb-org}
\emph{Instance-wise (triplet) batching} re-encodes the full history $\mathcal{H}$ for every $(u,v,y)$, driving dataset
size and I/O as $O(mL)$. \emph{Padding/bucketing by length} mitigates kernel divergence but still repeats user encoding.
\emph{History truncation / retrieval-first} shortens or selects subsequences, reducing cost but discarding information and
breaking end-to-end gradients through the full history. \emph{Embedding caching / clustering} compresses histories at the
cost of approximation error. In contrast, \textbf{Request-Level Batching (RLB)} is lossless and end-to-end: it preserves
the full $\mathcal{H}$, keeps the objective intact, and lowers the \emph{per-target} user-path complexity from $O(L)$ to
roughly $O(L/m)$. In practice we set $m{=}8$, yielding about $8\times$ lower user-side bandwidth and encoder compute,
higher GPU utilization, and longer feasible $L$.

\subsection{Extrapolation Details: Beta Sampling and Load Balancing}
\label{app:extrapolation-details}

\subsubsection{Stochastic length sampling with a Beta curriculum}
We sample a normalized length ratio $s\in(0,1)$ from a Beta distribution and map it to the training length:
\[
s \sim \text{Beta}(\alpha, \beta),
\]
\begin{equation}
L_{\text{train}}^{\text{raw}} = L_{\text{train}}^{\min} + s \cdot (L_{\text{train}}^{\max} - L_{\text{train}}^{\min}).
\label{eq:beta-scale}
\end{equation}
We use Beta because it can realize the empirically preferred \emph{U-shaped (bimodal)} distribution and it has only two
parameters, easing tuning. This sampling acts as a data curriculum; the end-to-end objective remains the same BCE loss
in Eq.~\eqref{eq:bce}.

\paragraph{Hardware alignment and discretization.}
To align with hardware acceleration requirements (e.g., tensor core alignment), we round $L_{\text{train}}^{\text{raw}}$
to the nearest multiple of 8 to obtain $L_{\text{train}}$.

\paragraph{Expectation constraint.}
Given a target average training length $L_{\text{train}}^{\text{avg}}$, we enforce
\[
\mathbb{E}[L_{\text{train}}^{\text{raw}}]
= L_{\text{train}}^{\min} + (L_{\text{train}}^{\max} - L_{\text{train}}^{\min}) \cdot \frac{\alpha}{\alpha + \beta}
= L_{\text{train}}^{\text{avg}},
\]
which implies
\begin{equation}
\beta = \alpha \cdot \frac{L_{\text{train}}^{\max} - L_{\text{train}}^{\text{avg}}}{L_{\text{train}}^{\text{avg}} - L_{\text{train}}^{\min}}.
\label{eq:beta}
\end{equation}
Thus $s$ is sampled from
\[
s \sim \text{Beta}\!\left(\alpha,\; \alpha \cdot \frac{L_{\text{train}}^{\max} - L_{\text{train}}^{\text{avg}}}{L_{\text{train}}^{\text{avg}} - L_{\text{train}}^{\min}}\right).
\]
The average length $L_{\text{train}}^{\text{avg}}$ directly controls sequence sparsity via
$\text{SS} = L_{\text{train}}^{\text{avg}} / L_{\text{train}}^{\max}$, while
$\rho_{\text{extra}}=L_{\text{infer}}/L_{\text{train}}^{\text{avg}}$.

\subsubsection{Batch-level load balancing and ragged target attention}
Variable-length sequences can cause workload imbalance, since the step time is dominated by the longest sequence in the
batch. We use a batch-level load-balancing operator with a fixed token budget of $B \cdot L_{\text{train}}^{\text{avg}}$
per batch:
\begin{enumerate}
    \item \textbf{Global length allocation}: stochastically truncate sequences so the total token count per batch stays
    close to $B \cdot L_{\text{train}}^{\text{avg}}$, improving workload balance while preserving the stochastic-length
    curriculum.
    \item \textbf{Sequence compaction}: compact/pack tokens to reduce padding; tokens from longer sequences are
    redistributed to shorter ones so the batch is processed near the average token budget.
\end{enumerate}
To efficiently process compacted variable-length sequences, we implement a ragged target-attention mechanism supported
by a high-throughput GEMM kernel~\cite{DBLP:conf/nips/DaoFERR22}. Instead of padding to the maximum length, we use an
auxiliary \texttt{index} tensor to mark segment boundaries: keys/values are flattened into 2D matrices, and each query
attends only to its corresponding segments in the flattened $K/V$.

\end{document}